%% file: Template.tex
\documentclass{article}
\input{preamble}

\title{OmniKVQuant: KV Cache Quantization for Omni-LLMs}

\name{
  Suho Yoo$^{1\:\ast}$ \quad
  Hyunjong Ok$^{2\:\ast}$ \quad
  Jongmin Choi$^1$ \quad
  Jihoo Jung$^1$ \quad
  Joon Son Chung$^1$
}
\address{
$^1$KAIST \quad$^2$POSTECH  \\
\thanks{\makebox[0pt][r]{$^\ast$}Equal contribution.}
}

\begin{document}
\maketitle

\setlength{\stripsep}{-45pt}
\begin{strip}
\centering
\includegraphics[width=\textwidth, height=0.25\textheight]{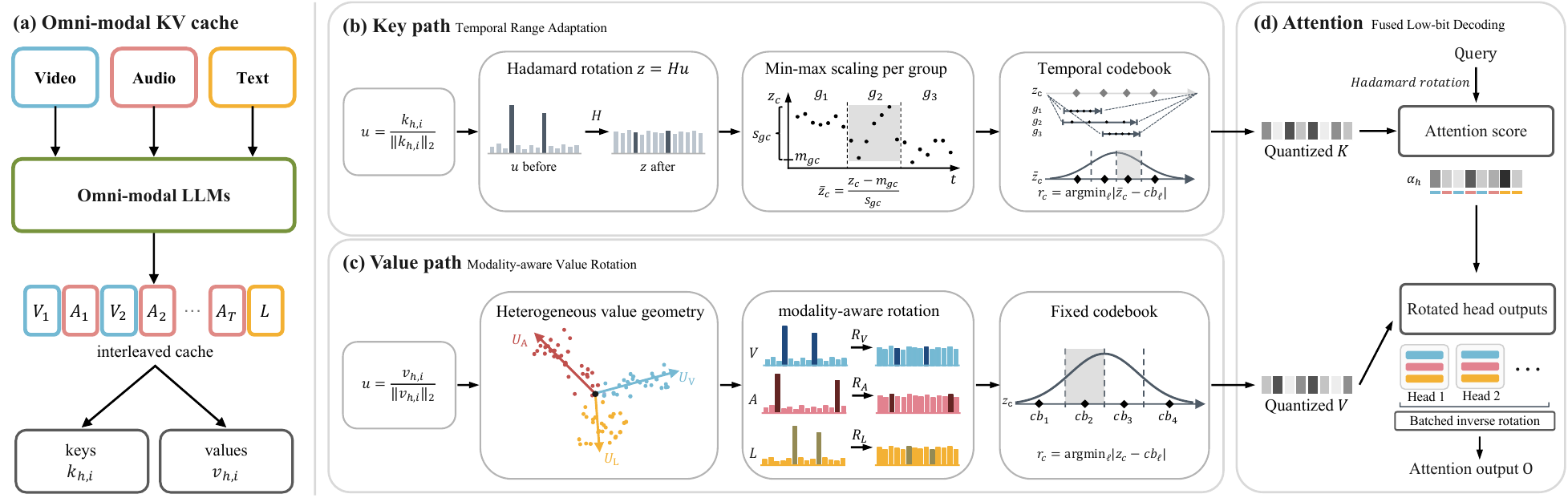}
\vspace{-20pt}
\captionof{figure}{\textbf{Overview of OmniKVQuant.}
We adapt key quantization ranges to local temporal shifts and apply modality-specific value rotations to capture heterogeneous geometry, enabling accurate KV caching and efficient attention through fused decoding.}
\label{fig:main}
\vspace{50pt}
\end{strip}

\begin{abstract}
% As Omni-modal large language models (Omni-LLMs) take in audio, video and text together, their KV cache keeps state for every stream at once, and its memory cost grows accordingly.
% KV cache quantization is the \emph{de~facto} approach to this problem in text-only LLMs, but its application to Omni-LLMs, whose cache interleaves audio, video and text, remains unexplored.
% In this paper, we analyze how TurboQuant, a representative rotation-based KV cache quantization method, behaves on multimodal caches, and identify two critical issues: \emph{temporal key drift}, where keys vary as the audio-visual stream goes on, and \emph{heterogeneous value geometry}, where each modality occupies a different part of the value space.
% To address these, we propose \textbf{\emph{OmniKVQuant}}, a training-free framework that (i) sets the key quantization range over each short window of the stream, following variation that a global one cannot; and (ii) rotates values separately per modality, along the directions the model actually reads.
% On Qwen2.5-Omni and Qwen3-Omni, OmniKVQuant enables 2-bit KV caches while substantially preserving performance across seven audio-visual benchmarks.
% We further provide a fused Triton decode kernel that unpacks the 2-bit cache during attention, so no dense FP16 cache is ever built.

As Omni-modal large language models (Omni-LLMs) take in audio, video and text together, their KV cache memory cost grows.
KV cache quantization is the \emph{de~facto} approach in text-only LLMs, but its application to Omni-LLMs remains unexplored.
In this paper, we analyze how TurboQuant, a representative rotation-based KV cache quantization method, behaves on multimodal caches and identify two critical issues: \emph{temporal key drift} and \emph{heterogeneous value geometry}.
To address these, we propose \textbf{\emph{OmniKVQuant}}, a training-free framework that (i) sets the key quantization range over each short window of the input stream; and (ii) rotates values separately per modality.
On Qwen2.5-Omni and Qwen3-Omni, OmniKVQuant enables 2-bit KV caches while substantially preserving performance across seven audio-visual benchmarks.
We further provide a fused Triton decode kernel that unpacks the 2-bit cache during attention, so no dense FP16 cache is ever built\footnote{\url{https://github.com/kaistmm/OmniKVQuant}}.

% \textbf{Code:} \href{https://github.com/kaistmm/OmniKVQuant}{\texttt{github.com/kaistmm/OmniKVQuant}}

\end{abstract}
\vspace{-5pt}

\input{sections/intro}
\input{sections/method}
\input{sections/experiments}

% \section{Conclusion}
% OmniKVQuant shows that low-bit omni-modal KV caches benefit from treating keys and values according to their distinct structure.
% Keys favor calibration-free local ranges, whereas values favor calibrated modality-specific rotations. Together, these design choices allow OmniKVQuant to consistently preserve performance across model scales and diverse audio-visual tasks under aggressive low-bit compression.
\section{Conclusion}
We presented OmniKVQuant, a training-free framework for KV cache quantization in Omni-LLMs.
OmniKVQuant uses local ranges for keys and modality-specific rotations for values, reflecting their different structures in the cache.
On Qwen2.5-Omni and Qwen3-Omni across seven audio-visual benchmarks, OmniKVQuant consistently outperforms TurboQuant under 2-bit KV cache quantization.
The gains hold across both multiple-choice and captioning tasks, showing that the same design carries across model scales and task types.
These results suggest that the structure of keys and values is important for compressing Omni-LLM KV caches.

% \vfill\pagebreak

\bibliographystyle{IEEEbib}
\bibliography{shortstrings,refs}

\end{document}

%% file: preamble.tex
\usepackage{spconf,amsmath,graphicx,hyperref}

\usepackage[utf8]{inputenc} % allow utf-8 input
\usepackage[T1]{fontenc}    % use 8-bit T1 fonts
\usepackage{url}            % simple URL typesetting
\usepackage{booktabs}       % professional-quality tables
\usepackage{amsfonts}       % blackboard math symbols
\usepackage{nicefrac}       % compact symbols for 1/2, etc.
\usepackage{microtype}      % microtypography
\usepackage{xcolor}         % colors

\usepackage[accsupp]{axessibility}  % Improves PDF readability for those with disabilities.

\usepackage{cite}
\usepackage{caption}
\usepackage{subcaption}
\usepackage{graphicx}
\usepackage{hyperref}
\usepackage{cleveref}

\usepackage{lipsum}
\usepackage{wrapfig}
\usepackage{array}
\usepackage{xcolor}
\usepackage{makecell}
\usepackage{enumitem}

\usepackage{algorithm}
\usepackage{algpseudocode}
\floatname{algorithm}{Algorithm}
\algrenewcommand\algorithmicrequire{\textbf{Require:}}
\algrenewcommand\algorithmicensure{\textbf{Ensure:}}

\usepackage{orcidlink}
\usepackage[dvipsnames]{xcolor}
\usepackage{booktabs}

\usepackage{pifont} % for cmark / xmark

\usepackage{colortbl}
\usepackage{multirow}
\usepackage{multicol}
\usepackage{listings} % Add code blocks
\usepackage{titletoc} % for table of contents of appendix only, should be BEFORE HYPERREF
\usepackage{fancyvrb}
\usepackage{tcolorbox} % Prompts in appendix
\usepackage[accsupp]{axessibility}  % Improves PDF readability for those with disabilities.
\usepackage[subtle]{savetrees} % subtle | moderate | extreme

\usepackage{cuted}
\usepackage{caption}

\newcommand{\subpara}[1]{%
  \vspace{1mm}%
  \noindent\textbf{#1}%
}

\definecolor{lightgray}{rgb}{0.83, 0.83, 0.83}
\definecolor{Gray}{gray}{0.6}
\definecolor{aliceblue}{rgb}{0.94, 0.97, 1.0}
\definecolor{mistyrose}{rgb}{1.0, 0.89, 0.88}
\definecolor{backcolour}{rgb}{0.95,0.95,0.92}

\newcommand{\appendixref}[2]{%
  \if\sepappendix1%
    #1% If separate appendix
  \else%
    #2% If together
  \fi%
}

\usepackage{cleveref}

\crefname{equation}{Eq.}{Eqs.}
\Crefname{equation}{Equation}{Equations}

\crefname{figure}{Fig.}{Figs.}
\Crefname{figure}{Figure}{Figures}

\crefname{table}{Tab.}{Tabs.}
\Crefname{table}{Table}{Tables}

\crefname{section}{Sec.}{Secs.}
\Crefname{section}{Section}{Sections}

\crefname{algorithm}{Alg.}{Algs.}
\Crefname{algorithm}{Algorithm}{Algorithms}

\usepackage{etoolbox}
\AtBeginDocument{%
  \setlength{\abovedisplayskip}{3pt}%
  \setlength{\belowdisplayskip}{3pt}%
  \setlength{\abovedisplayshortskip}{2pt}%
  \setlength{\belowdisplayshortskip}{2pt}%
}

%% file: sections/intro.tex
\section{Introduction}
\label{sec:intro}

Recent Omni-modal large language models (Omni-LLMs) perceive audio, vision and text within a single model, and perform strongly on real-world audio-visual understanding~\cite{xu2025qwen25omnitechnicalreport, xu2025qwen3, hurst2024gpt, jung2026avcd, yoo2026nature}. However, the three streams do not come for free. The model holds all three at once, a single minute of audio-visual input can exceed 10,000 tokens, each token leaving behind a key and a value, and the cache soon takes more memory than the weights themselves.
Keeping it small without giving up accuracy is therefore central to deploying Omni-LLMs~\cite{tao2026omnizip, yoo2026out}.

The usual way to reduce it is KV cache quantization~\cite{hooper2024kvquant}, storing each coordinate in fewer bits than the full precision.
This is hard because a few coordinates of the cache are much larger  than the rest~\cite{hooper2024kvquant, pmlr-v235-liu24bz, xiao2023smoothquant}. The bits have to reach the large ones, so the small ones, which are most of the cache, all come out looking alike.
TurboQuant~\cite{zandieh2026turboquant} addresses this by rotating each KV cache vector before storing it, mixing the large dimensions into the others so that no single one is left to set the scale.

% \begin{figure}[t]
% \centering
%   \begin{subfigure}{0.49\linewidth}
%     \centering
%     \includegraphics[width=\linewidth]{figures/fig1_a_3b.pdf}
%     \subcaption{Qwen2.5-Omni$_{3\mathrm{B}}$}\label{fig:radar-a}
%   \end{subfigure}\hfill
%   \begin{subfigure}{0.49\linewidth}
%     \centering
%     \includegraphics[width=\linewidth]{figures/fig1_b_30b.pdf}
%     \subcaption{Qwen3-Omni$_{30\mathrm{B}}$}\label{fig:radar-b}
%   \end{subfigure}
%   \caption{\textbf{OmniTurbo remains consistently close to the uncompressed FP16 cache across benchmarks, unlike the modality-blind baseline.} Its stability across modalities enables reliable 2-bit KV cache compression in practice.}
%   \label{fig:radar}
%   \vspace{-15pt}
% \end{figure}

This approach has since been applied to other domains, to protein sequences~\cite{hu2026turboesm} and to audio, video and text models taken one at a time~\cite{boss2026octopus}, but never to a cache that holds more than one modality.
MASQuant~\cite{hu2026masquant} does account for audio, video and text jointly, finding that a shared treatment follows the modality with the largest activations, but it does so for the weights and the activations, not the KV cache.
No existing method quantizes the KV cache for Omni-LLMs.

In this paper, we take TurboQuant as our starting point and ask where its assumptions fail when the cache is written by three streams rather than one. 
The first failure is \emph{temporal key drift}. Key distributions differ across modalities and vary even more substantially across short temporal windows. TurboQuant fixes its quantization scale in advance without measuring the cache it writes, leaving it unable to track these shifts as the multimodal stream goes on.
The second is \emph{heterogeneous value geometry}. Each modality occupies a different part of the value space, so one rotation shared by all three spreads them less effectively than a rotation chosen for each.
To address this, we introduce \textbf{\emph{OmniKVQuant}}, the first KV cache quantization method for Omni-LLMs. OmniKVQuant adapts the key codebook scale to the sequence as it progresses, following variation a fixed scale misses (\cref{fig:main}b). For the values, it calibrates a separate rotation for each modality, shaped to the geometry each one actually sits in (\cref{fig:main}c).

Experiments on Qwen2.5-Omni and Qwen3-Omni show that OmniKVQuant consistently outperforms TurboQuant. Across seven audio-visual benchmarks on Qwen2.5-Omni, OmniKVQuant retains 98.1\% of FP16 performance, even at 2-bit precision, demonstrating the importance of accounting for the KV cache structure of Omni-LLMs in low-bit quantization.

% We further provide a fused Triton decode kernel following FlashAttention~\cite{dao2022flashattentionfastmemoryefficientexact,dao2023flashattention2fasterattentionbetter}. It rotates one query per head instead of every cached key and batches inverse value rotations across head outputs, reducing redundant computation and launch overhead without reconstructing a dense FP16 cache.
We further provide a fused Triton decode kernel using the computation order of FlashAttention~\cite{dao2022flashattentionfastmemoryefficientexact,dao2023flashattention2fasterattentionbetter}. Our kernel rotates one query per head instead of inverse-rotating every cached key and batches inverse value rotations across head outputs (\cref{fig:main}d), reducing redundant rotation work and kernel launch overhead without reconstructing a dense FP16 cache.

% Our contributions are threefold. First, we analyze how TurboQuant behaves on multimodal KV caches and identify where its assumptions do not hold for interleaved audio, video, and text. Second, based on this analysis, we develop OmniKVQuant, a training-free KV cache quantization method that adapts key quantization to temporal statistics and value rotation to modality-specific geometry. Third, we provide a fused Triton decoding kernel that directly consumes the 2-bit KV cache, avoiding reconstruction of a dense FP16 cache during attention and enabling efficient low-bit inference.

%% file: sections/method.tex
\section{Method}

\input{tables/temporal_key}

\subsection{Preliminaries}

\subpara{Background on Omni-LLMs.}
Omni-LLMs jointly process visual, audio, and text inputs within a shared LLM decoder.
For a video with its audio stream, the input is organized into $T$ temporal chunks, where each chunk contains visual tokens $V_t$ followed by audio tokens $A_t$.
Together with textual tokens $L$, the multimodal sequence is written as
\[
X
=
[V_1,A_1,V_2,A_2,\ldots,V_T,A_T,L]
\in\mathbb{R}^{N\times D},
\]
where $N$ denotes the sequence length and $D$ the hidden dimension.
At each transformer layer and attention head $h$, the hidden states are projected into queries, keys, and values as
$Q_h=XW_h^Q$,
$K_h=XW_h^K$, and
$V_h=XW_h^V$.
RoPE~\cite{su2024roformer} then applies position-dependent rotations to the query and key representations,
while the value representations remain unrotated.
Attention is computed as
\begin{equation}
\alpha_h
=
\operatorname{Softmax}
\left(
\frac{
Q_h
K_h^{\top}
}{
\sqrt{d}
}
\right),
\qquad
O_h
=
\alpha_hV_h,
\label{eq:omni_attn}
\end{equation}
where $d$ denotes the head dimension.
During autoregressive decoding, the resulting key and value representations are stored in the KV cache and reused for subsequent tokens.

\subpara{TurboQuant.}
TurboQuant~\cite{zandieh2026turboquant} is applied independently to each
$k_{h,i}\in K_h$ and $v_{h,i}\in V_h$ stored in the KV cache.
For either $x\in\{k_{h,i},v_{h,i}\}$, it first separates the magnitude
from the unit direction and applies a randomized Hadamard transform,
\begin{equation}
u=\frac{x}{\lVert x\rVert_2},
\qquad
z=Hu,
\label{eq:tq_hadamard}
\end{equation}
The rotation makes each coordinate $z_c$ approximately follow
$\mathcal{N}(0,1/d)$ in high dimensions.
A fixed codebook $cb=(cb_1,\ldots,cb_{2^b})$ is then constructed
from this Gaussian reference distribution, and the quantized vector is
reconstructed as $\hat{x}$ through inverse rotation.
\begin{equation}
r_c=\arg\min_\ell |z_c-cb_\ell|,
\qquad
\hat{x}
=
\lVert x\rVert_2 H^\top cb[r].
\label{eq:tq_codebook}
\end{equation}
In this way, TurboQuant spreads a few large dimensions through random rotation and brings keys and values closer to a Gaussian reference distribution, allowing a fixed codebook to be shared across the KV cache.

%k는 hadamard를 해줘서 전체적으로는 가우시안인데, 이게 omni에서는 여러 modality + 시간축 변화가 있는데 value와 달리 key는 특히 분산 이 더 작음 (modality 내, modality 간 관찰 + key, value 관찰). 그래서 localize가 된다(beta 분포에서 key, value가 로컬에서 얼마나 벗어나는지?) + rope 영향. 부분적으로 가우시안에서 많이 벗어나더라~ 그래서 g32를 해주는거고. value는 그렇진 않음 근데? value는 representation자체가 직접 head output에 반영됨(key는 query에 따라 attn에 변동이 큰 반면.) -> 근데 value geometry 관찰해보면 modality 별로 나뉘어 있고 이걸 attn weighted로 회전을 fitting 해놓으면 더 괜찮더라~(value or attn head PCA)

\begin{figure}[t]
\centering
\vspace{-0.3em}
\includegraphics[width=\linewidth]{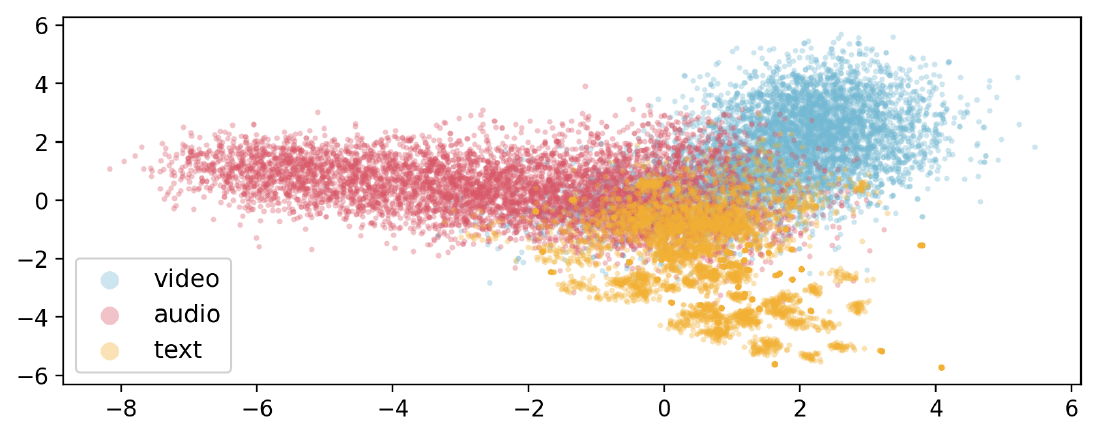}
\vspace{-2.2em}
\caption{\textbf{Heterogeneous value geometry.}
Value representations form distinct modality-specific distributions, with text, audio, and video occupying different regions of the feature space.}
\label{fig:pca}
\vspace{-1.4em}
\end{figure}

\subsection{Observations}
We analyze Qwen2.5-Omni-3B~\cite{xu2025qwen25omnitechnicalreport} on WorldSense~\cite{hong2025worldsense}.

\input{tables/main_onecolumn}

\subpara{Temporal key drift.}
We measure the retained variance relative to the Gaussian distribution, together with the Wasserstein-1 (W1) distance and Kolmogorov--Smirnov (KS) statistic.
Although the Hadamard rotation is designed to align coordinates with a Gaussian distribution, cached keys exhibit a substantial deviation from this target.
As shown in \cref{tab:distribution_g32}, key coordinates retain only $30.40\%$ of the Gaussian variance, with W1 and KS distances of $0.778$ and $0.449$, respectively.
This mismatch persists across modalities and is more pronounced for audio and video than for text.
In contrast, value representations remain considerably closer to the Gaussian distribution.
These results indicate that key distributions are not fully stabilized by the global Hadamard rotation and instead exhibit local variation across the cache, motivating our temporal range adaptation.

\subpara{Heterogeneous value geometry.}
Values directly contribute to the attention output, making their representation geometry important.
We visualize value representations across modalities using PCA in \cref{fig:pca}.
Visual, audio, and text values occupy distinct regions of the representation space, revealing modality-specific structures.
This suggests that a modality-aware correction of the Hadamard rotation can better adapt to heterogeneous value structures.

\subsection{OmniKVQuant}

\subpara{Temporal range adaptation.}
TurboQuant directly quantizes the rotated key direction $z$ using the fixed codebook in~\cref{eq:tq_codebook}.
However, we observe that the range of $z$ drifts across temporal positions, so a single shared codebook does not consistently match different cache regions.
We therefore min-max scale $z$ within a small group of consecutive tokens before codebook assignment.
For each group $g$ and coordinate $c$, we compute its minimum $m_{gc}$ and scale $s_{gc}$, and replace $z_{c}$ in~\cref{eq:tq_codebook} with
\begin{equation}
\bar{z}_{c}
=
\frac{z_{c}-m_{gc}}{s_{gc}}.
\label{eq:key_norm}
\end{equation}
The same fixed codebook is then applied to $\bar{z}$, while $m_{gc}$ and $s_{gc}$ are stored to restore the original range at reconstruction.
This allows the shared codebook to adapt to local temporal variation without modifying the Hadamard basis.
In our code, we compute attention scores in the rotated space by rotating a single query per head instead of inverse rotating every cached key, avoiding rotation costs that grow with cache length.

\subpara{Modality-aware value rotation.}
For values, the mismatch arises from distinct representation geometries across modalities that a shared rotation does not adequately capture.
We therefore calibrate a separate modality-specific basis $U_m$ for each modality
$m\in\{V,A,L\}$.

Rather than treating all value directions equally, we weight each value according to how strongly it contributes to the attention output.
Specifically,
\begin{equation}
w_j=\sum_i \alpha_{i,j}^{2},
\qquad
C_m=\sum_j w_j\,v_j^{\top}v_j,
\label{eq:value_weight}
\end{equation}
where $C_m$ is computed separately for each modality.
We obtain $U_m$ from the eigen-decomposition of $C_m$, so that each basis captures the value directions that contribute most strongly to the model output.
We estimate these modality-specific bases using calibration data from VGGSound~\cite{chen2020vggsound}.

We replace the Hadamard rotation $z=Hu$ in~\cref{eq:tq_hadamard} by
\begin{equation}
z=H U_m^{\top}u = R_m u,
\qquad
\hat{x}
=
\lVert x\rVert_2 R_m^{\top}cb[r],
\label{eq:modality_rotation}
\end{equation}
where $r$ denotes the quantization indices in~\cref{eq:tq_codebook}.
In practice, our Triton kernel applies the inverse rotation after aggregation within each attention head, rather than to each cached value.

%% file: tables/temporal_key.tex
\begin{table}[t]
\centering
\caption{\textbf{Distribution analysis over 32-token windows.}
Keys exhibit substantially lower local variance than a Gaussian distribution, especially for audio and video tokens.}
\vspace{-10pt}
\label{tab:distribution_g32}
\setlength{\tabcolsep}{20pt}
\small
\resizebox{\linewidth}{!}{%
\begin{tabular}{lccc}
\toprule
Distribution & Var.$\uparrow$ & W1$\downarrow$ & KS$\downarrow$ \\
\midrule
\rowcolor{gray!8}
Gaussian & 96.88\% & 0.214 & 0.149 \\
\midrule
Key                  & 30.40\% & 0.778 & 0.449 \\
\quad $\hookrightarrow$ Text  & 35.84\% & 0.738 & 0.421 \\
\quad $\hookrightarrow$ Audio & 28.67\% & 0.788 & 0.454 \\
\quad $\hookrightarrow$ Video & 28.70\% & 0.794 & 0.461 \\
\midrule
Value                & 73.06\% & 0.468 & 0.267 \\
\bottomrule
\end{tabular}%
}
\vspace{-1.1em}
\end{table}

%% file: tables/main_onecolumn.tex
\begin{table*}[t]
\centering
\caption{\textbf{Main results.} All compressed methods use K2/V2 KV cache quantization.
\emph{Avg.} normalizes each benchmark score to its FP16 counterpart before averaging. The FP16 baseline is shown in \textcolor{gray}{gray}.}
\vspace{-10pt}
\label{tab:main}
\setlength{\tabcolsep}{10pt}\small
\resizebox{\linewidth}{!}{%
\begin{tabular}{lccccc|cccc}
\toprule
& \multicolumn{5}{c}{\textit{Multiple choice} (Acc.)}
& \multicolumn{4}{c}{\textit{Captioning} (LLM-judge)} \\
\cmidrule(lr){2-6}\cmidrule(lr){7-10}
Method
& World.$\uparrow$ & Daily.$\uparrow$ & MME$\uparrow$ & OmniVid.$\uparrow$ & Avg.$\uparrow$(\%)
& UGC$\uparrow$ & DiaDem$\uparrow$ & SAL.2$\downarrow$ & Avg.$\uparrow$(\%) \\
\midrule
\multicolumn{10}{c}{\textit{Qwen2.5-Omni-3B} (Dense)} \\
\midrule
\rowcolor{gray!8}
FP16
& 47.6 & 57.5 & 75.8 & 44.0 & 100.0
& 57.2 & 18.4 & 83.5 & 100.0 \\
TurboQuant
& 45.6 & 56.1 & 73.6 & 42.8 & 96.9
& 52.5 & 16.8 & 90.5 & 91.7 \\
\rowcolor{aliceblue}
\textbf{OmniKVQuant}
& \textbf{47.8} & \textbf{54.8} & \textbf{75.3} & \textbf{44.0} & \textbf{98.8}
& \textbf{55.9} & \textbf{17.8} & \textbf{85.4} & \textbf{97.3} \\
% OmniKVQuant (FP16 meta.)
% & 47.1 & 55.3 & 74.5 & 44.6 & 98.7
% & 56.7 & 15.0 & 84.3 & 93.1 \\
\midrule
\multicolumn{10}{c}{\textit{Qwen3-Omni-30B-A3B} (MoE)} \\
\midrule
\rowcolor{gray!8}
FP16
& 53.4 & 69.7 & 85.3 & 48.8 & 100.0
& 71.8 & 23.0 & 67.9 & 100.0 \\
TurboQuant
& 33.3 & 31.1 & 56.7 & 29.5 & 58.5
& 41.8 & 1.3 & 101.9 & 43.6 \\
\rowcolor{aliceblue}
\textbf{OmniKVQuant}
& \textbf{48.1} & \textbf{62.2} & \textbf{80.5} & \textbf{43.4} & \textbf{90.6}
& \textbf{66.2} & \textbf{17.6} & \textbf{83.1} & \textbf{83.4} \\
% OmniKVQuant (FP16 meta.)
% & 49.1 & 61.9 & 80.5 & 50.6 & 94.7
% & 67.6 & 19.0 & 79.8 & 87.3 \\
\bottomrule
\end{tabular}%
}
\vspace{-1em}
\end{table*}

%% file: sections/experiments.tex
\section{Experiments}

\input{tables/mse}

\subsection{Setup}

\subpara{Datasets and metrics.} We evaluate on seven audio-visual benchmarks. 
Multiple-choice tasks include WorldSense~\cite{hong2025worldsense} (World.), DailyOmni~\cite{zhou2025daily} (Daily.), Video-MME~\cite{fu2025video} (MME), and OmniVideoBench~\cite{li2025omnivideobench} (OmniVid.), all scored by accuracy. For captioning, we use UGC-VideoCap~\cite{wu2025ugcvideocap} (UGC.), DiaDemBench~\cite{chen2026diadem} (DiaDem.), and video-SALMONN2~\cite{tang2025video} (SAL.2), evaluated by Qwen3.8-27B~\cite{qwen38} as an LLM judge.
Avg. normalizes each benchmark to the FP16 (100) and averages across benchmarks, using the inverse ratio for SAL.2.
Due to FP16 memory limits, we evaluate samples up to one minute.

\subpara{Baselines.}
We compare three cache configurations.
FP16 is the uncompressed cache.
TurboQuant~\cite{zandieh2026turboquant} is implemented following vLLM~\footnote{\url{https://github.com/vllm-project/vllm}}, with TurboQuant applied to both keys and values.
In our experiments, both methods quantize keys and values to 2 bits (K2/V2).
Except for the proposed codec components of OmniKVQuant, both methods use the same inference pipeline.

\subpara{Implementation details.} 
We use Qwen2.5-Omni-3B~\cite{xu2025qwen25omnitechnicalreport} and Qwen3-Omni-30B-A3B-Instruct~\cite{xu2025qwen3} under greedy decoding on a single NVIDIA A100 and H200 GPU, respectively.
We use 32-token groups $g$ for key range estimation, with MXFP4 metadata by default, and calibrate value rotations once per model on 309 VGGSound training clips, one per class.
For multiple-choice evaluation, we prefill all but the final prompt token and decode one token with the quantized KV cache. For captioning, OmniKVQuant keys are quantized every 128 tokens in groups of 32, while OmniKVQuant values and TurboQuant keys/values use a 128-token sliding FP16 window before quantization.

\subsection{Experimental Results}

\cref{tab:main} reports results across four multiple-choice and three captioning benchmarks.
On Qwen2.5-Omni, OmniKVQuant retains 98.8\% and 97.3\% of FP16 performance on multiple-choice and captioning, respectively, substantially outperforming TurboQuant at the same K2/V2 budget.
The gap widens on Qwen3-Omni, where TurboQuant drops to 58.5\% and 43.6\%, while OmniKVQuant preserves 90.6\% and 83.4\%.
These results show that quantization tailored to the KV cache structure of Omni-LLMs is critical for preserving performance.

\subsection{Analysis}

Unless otherwise noted, all analyses use Qwen2.5-Omni-3B.

\subpara{Error comparison.}
\cref{tab:quant_comparison} compares the codecs directly on World., with both keys and values quantized.
\emph{Recon. MSE} measures key cache reconstruction error, \emph{Attn.\ KL} the resulting attention-distribution shift, and \emph{Output MSE} the error in the attention output.
OmniKVQuant consistently reduces all three errors at a matched bit rate. For an exact K3-budget comparison, we use FP16 minimum $m_{gc}$ and scale $s_{gc}$ metadata over 32-token groups, which adds one bit per element yet still yields lower error than 3-bit key quantization.

\input{tables/ablation}
\input{tables/bits_ablation}

\subpara{Ablation studies.}
\cref{tab:ablation} provides a component-wise ablation tracing the progression from TurboQuant to our full method. The results show that the local key range is particularly important, while modality-specific rotation further stabilizes performance across benchmarks. \cref{tab:ladder} examines different compression rates and shows that our method consistently outperforms TurboQuant across rate settings.

\begin{figure}[t]
\centering
\includegraphics[width=\linewidth]{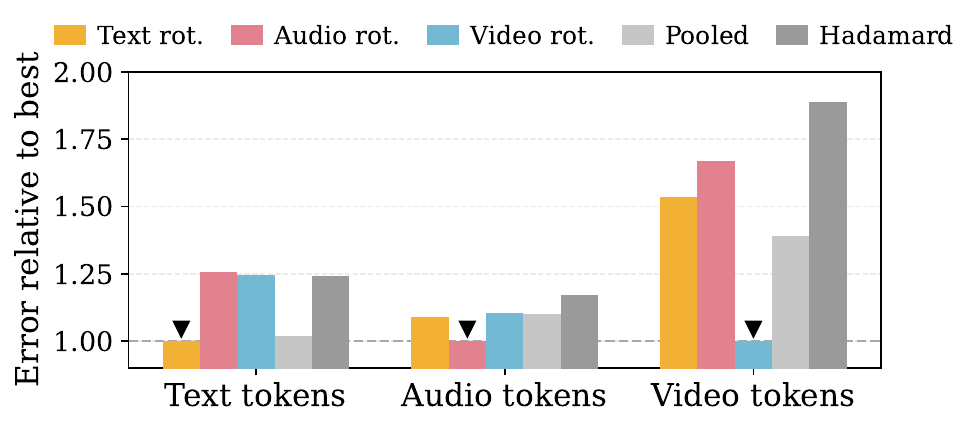}
\vspace{-20pt}
\caption{\textbf{Value quantization error across rotation choices.}
Using the rotation matched to each token modality ($\blacktriangledown$) yields the lowest error in all three groups, while pooled and shared rotations are consistently worse.}
\label{fig:vtransfer}
\vspace{-0.2em}
\end{figure}

\subpara{Value rotation.}
\cref{fig:vtransfer} examines how value rotations transfer across modalities on World.
For text, audio, and video tokens, the rotation calibrated on the same modality consistently gives the lowest quantization error, while rotations transferred from other modalities or calibrated on pooled data perform worse.
The same pattern holds across all three modalities, showing that their value representations favor different quantization bases rather than a single shared rotation.

\input{tables/k_grouping}

\subpara{Key adaptation.}
With $g=32$, \cref{tab:kgroup} separates modality identity from temporal locality on World.
\emph{Range} measures the range covered by each group.
Global shuffling gives the largest key error, while shuffling only within each modality recovers part of the loss.
Keeping tokens temporally contiguous further reduces both key range and downstream quantization error.
This shows that modality identity alone is insufficient: key statistics continue to vary locally within each modality.

%K: 모달리티 정보만으로 부족 → 매 g32에서 min/span 실측
%V: 모달리티별 고차원 basis가 다름 → offline rotation 3개

%% file: tables/mse.tex
\begin{table}[t]
\centering
\caption{\textbf{Quantization error at comparable bit budgets.}
Our method consistently reduces error compared with TurboQuant.}
\vspace{-10pt}
\label{tab:quant_comparison}
\setlength{\tabcolsep}{2pt}
\small
\resizebox{\linewidth}{!}{%
\begin{tabular}{lcccc}
\toprule
Method & Recon. MSE$\downarrow$ & Attn. KL$\downarrow$ & Output MSE$\downarrow$ & Bits/elem.$\downarrow$ \\
\midrule
TurboQuant (K2/V2) & 1.2939 & 0.6739 & 0.0345 & 2.125 \\
\rowcolor{aliceblue}
\textbf{Ours (MXFP4 meta.)} & \textbf{0.1833} & \textbf{0.1510} & \textbf{0.0148} & 2.258 \\
\midrule
TurboQuant (K3/V2) & 0.3918 & 0.2436 & 0.0159 & 2.625 \\
\rowcolor{aliceblue}
\textbf{Ours (FP16 meta.)} & \textbf{0.1495} & \textbf{0.1167} & \textbf{0.0134} & 2.625 \\
\bottomrule
\end{tabular}%
}
\vspace{-15pt}
\end{table}

%% file: tables/ablation.tex
% \begin{table}[t]
% \centering
% \captionsetup{width=1\columnwidth}
% \caption{\textbf{Ablation from TurboQuant to our full method.} Local key ranges drive most of the gain, while modality-specific rotation further stabilizes performance.}
% \vspace{-10pt}
% \label{tab:ablation}
% \setlength{\tabcolsep}{10pt}\small
% \resizebox{\linewidth}{!}{%
% \begin{tabular}{lcccc}
% \toprule
% Method & Local-K & Rot. & MCQ Avg.$\uparrow$(\%) & Cap.$\uparrow$ \\
% \midrule
% \multicolumn{5}{c}{\textit{Qwen2.5-Omni$_{3\mathrm{B}}$} (dense)} \\
% \midrule
% TurboQuant & & & 96.9 & 1.68 \\
% % + Local key ranges
% & \checkmark & & 98.5 & \textbf{1.98} \\
% % + Modality value rot. & & \checkmark & 96.7 & 1.64 \\
% \rowcolor{aliceblue}
% % \textbf{Ours} 
% & \checkmark & \checkmark & \textbf{98.8} & 1.93 \\
% \midrule
% \multicolumn{5}{c}{\textit{Qwen3-Omni-30B}$_{A3\mathrm{B}}$ (MoE)} \\
% \midrule
% TurboQuant & & & 58.5 & 1.29 \\
% % + Local key ranges 
% & \checkmark & & 91.2 & 2.14 \\
% % + Modality value rot. & & \checkmark & 55.7 & 1.28 \\
% \rowcolor{aliceblue}
% % \textbf{Ours} 
% & \checkmark & \checkmark & \textbf{93.5} & \textbf{2.14} \\
% \bottomrule
% \end{tabular}%
% }
% \vspace{-5pt}
% \end{table}

\begin{table}[t]
\centering
\captionsetup{width=1\columnwidth}
\caption{\textbf{Component ablation.}
Local key ranges provide the largest improvement over TurboQuant, while modality-specific value rotation yields additional gains across both model scales.}
\vspace{-10pt}
\label{tab:ablation}
\setlength{\tabcolsep}{7pt}\small
\resizebox{\linewidth}{!}{%
\begin{tabular}{lcccc}
\toprule
Method & Key & Value & MCQ$\uparrow$ & Cap.$\uparrow$ \\
\midrule
\multicolumn{5}{c}{\textit{Qwen2.5-Omni-3B} (Dense)} \\
\midrule
TurboQuant & & & 96.9 & 91.7 \\
+ Local key ranges
& \checkmark & & 98.5 & 95.2 \\
% + Modality value rot. & & \checkmark & 96.7 & 93.6 \\
\rowcolor{aliceblue}
+ Modality value rot. (Ours)
& \checkmark & \checkmark & \textbf{98.8} & \textbf{97.3} \\
\midrule
\multicolumn{5}{c}{\textit{Qwen3-Omni-30B-A3B} (MoE)} \\
\midrule
TurboQuant & & & 58.5 & 43.6 \\
+ Local key ranges
& \checkmark & & \textbf{91.2} & 78.1 \\
% + Modality value rot. & & \checkmark & 55.6 & 44.1 \\
\rowcolor{aliceblue}
+ Modality value rot. (Ours)
& \checkmark & \checkmark & 90.6 & \textbf{83.4} \\
\bottomrule
\end{tabular}%
}
\vspace{-0.3em}
\end{table}

% 밑에가 벤치 2개씩 
% 밑에가 벤치 2개씩 
% 밑에가 벤치 2개씩 
% \begin{table}[t]
% \centering
% \captionsetup{width=1\columnwidth}
% \caption{}
% \vspace{-10pt}
% \label{tab:ablation}
% \setlength{\tabcolsep}{5pt}\footnotesize
% \resizebox{\linewidth}{!}{%
% \begin{tabular}{lcc|cc}
% \toprule
% & \multicolumn{2}{c|}{\textit{Multiple choice}} & \multicolumn{2}{c}{\textit{Captioning}} \\
% \cmidrule(lr){2-3}\cmidrule(lr){4-5}
% Method & MME & OmniVid. & DiaDem & SAL.2$\downarrow$ \\
% \midrule
% \multicolumn{5}{c}{\textit{Qwen2.5-Omni-3B} (Dense)} \\
% \midrule
% TurboQuant & 73.6 & 42.8 & 16.8 & 90.5 \\
% + Local key ranges & \textbf{76.2} & 42.2 & 16.8 & 86.4 \\
% \rowcolor{aliceblue}
% + Modality value rot. (Ours) & 75.3 & \textbf{44.0} & \textbf{17.8} & \textbf{85.4} \\
% \midrule
% \multicolumn{5}{c}{\textit{Qwen3-Omni-30B-A3B} (MoE)} \\
% \midrule
% TurboQuant & 56.7 & 29.5 & 1.3 & 101.9 \\
% + Local key ranges & 79.7 & \textbf{45.8} & 14.8 & 85.4 \\
% \rowcolor{aliceblue}
% + Modality value rot. (Ours) & \textbf{80.5} & 43.4 & \textbf{17.6} & \textbf{83.1} \\
% \bottomrule
% \end{tabular}%
% }
% \vspace{-0.3em}
% \end{table}

%% file: tables/bits_ablation.tex
\begin{table}[t]
\centering
\caption{\textbf{Multiple-choice performance at different bit widths.}
OmniKVQuant consistently outperforms TurboQuant from 2-bit to 4-bit quantization, demonstrating generalizes well across quantization levels.}
\vspace{-0.9em}
\label{tab:ladder}
\setlength{\tabcolsep}{12pt}\small
\resizebox{\linewidth}{!}{%
\begin{tabular}{lccc}
\toprule
Method & K2V2$\uparrow$(\%) & K3V3$\uparrow$(\%) & K4V4$\uparrow$(\%) \\
\midrule
TurboQuant & 96.9 & 98.3 & 100.4 \\
\rowcolor{aliceblue}
\textbf{Ours} & \textbf{98.8} & \textbf{99.6} & \textbf{101.5} \\
\bottomrule
\end{tabular}%
}
\vspace{-10pt}
\end{table}

%% file: tables/k_grouping.tex
\begin{table}[t]
\centering
\caption{\textbf{Effect of temporal adaptation on key quantization.}
Modality information alone is insufficient; preserving local temporal structure further reduces key quantization error.}
\vspace{-0.425em}
\label{tab:kgroup}
\setlength{\tabcolsep}{2pt}\small
\resizebox{\linewidth}{!}{%
\begin{tabular}{lcccc}
\toprule
Grouping & Range$\downarrow$ & Recon. MSE$\downarrow$ & Attn.\ KL$\downarrow$ & Output MSE$\downarrow$ \\
\midrule
Shuffled globally & 0.2318 & 0.2716 & 0.2197 & 0.0188 \\
Shuffled within modality & 0.2251 & 0.2579 & 0.2054 & 0.0180 \\
\rowcolor{aliceblue}
Temporally contiguous & \textbf{0.1788} & \textbf{0.1833} & \textbf{0.1510} & \textbf{0.0148} \\
\bottomrule
\end{tabular}%
}
\vspace{-10pt}
\end{table}